# A Survey on Applications of Reinforcement Learning in Spatial Resource Allocation

Di Zhang[1], Moyang Wang[1], Joseph Mango[2], Xiang Li[1,3,4 *], Xianrui Xu[5]

[1]Key Laboratory of Geographic Information Science (Ministry of Education) and School of Geographic Sciences, East China Normal University, Shanghai 200241, People's Republic of China

[2]Department of Transportation and Geotechnical Engineering, University of Dar es Salaam, Dar es salaam, Tanzania

[3]Shanghai Key Lab for Urban Ecological Processes and Eco-Restoration, East China Normal University, Shanghai 200241, People's Republic of China

[4]Key Laboratory of Spatial-Temporal Big Data Analysis and Application of Natural Resources in Megacities (Ministry of Natural Resources), East China Normal University, Shanghai 200241, People's Republic of China

[5]School of Economics and Management, Shanghai University of Sport, Shanghai 200438, People's Republic of China

## Abstract

The challenge of spatial resource allocation is pervasive across various domains such as transportation, industry, and daily life. As the scale of real-world issues continues to expand and demands for real-time solutions increase, traditional algorithms face significant computational pressures, struggling to achieve optimal efficiency and real-time capabilities. In recent years, with the escalating computational power of computers, the remarkable achievements of reinforcement learning in domains like Go and robotics have demonstrated its robust learning and sequential decision-making capabilities. Given these advancements, there has been a surge in novel methods employing reinforcement learning to tackle spatial resource allocation problems. These methods exhibit advantages such as rapid solution convergence and strong model generalization abilities, offering a new perspective on resolving spatial resource allocation problems. Therefore, this paper aims to summarize and review recent theoretical methods and applied research utilizing reinforcement learning to address spatial resource allocation problems. It provides a summary and comprehensive overview of its fundamental principles, related methodologies, and applied research. Additionally, it highlights several unresolved issues that urgently require attention in this direction for the future.

# 1 Introduction

Conventional methods used to solve spatial resource allocation problems (Eyles et al., 1982; Long et al., 2018) is that the planners make effective allocations of the determined spatial resources with few reasonable parameters. Scientific and rational spatial resource allocation parameters can effectively reduce resource operation costs and improve its utilisation and output efficiency. Particularly, in the problem of allocating a freight warehouse, for example, the scientific and reasonable parameters can improve utilisation rates of the warehouse space, speed up the flows of goods and reduce inventory rates to a certain extent (Azadivar, 1989; Sanei et al., 2011). The successes of such goals depend on the effective planning of resource allocation. For enterprise planning, the first thing to consider is the scientific and reasonable layouts of the factory area to ensure their logistics and information flows run smoothly (Benjaafar et al., 2002; Naranje et al., 2019). The foremost things to consider for public-service facilities planning are their construction and maintenance costs and service coverages to reduce long distances and waiting time for users to obtain good services (Suchman, 1968; Wang et al., 2021). Overall, all such demands for effective resource allocation planning make the spatial resource allocation problem as a typical non-deterministic polynomial-time hardness (Dorit S. Hochba) problem (Dorit S Hochba, 1997) with complex decision-making environment challenges, specification description, interleaved spatial structure, and scale inhibitory effect on the search for high-quality solutions.

Numerous studies on the computational challenges of spatial resource allocation optimisation problems have been done using precise and heuristic methods. These precise methods, including the location-allocation model and quadratic programming, have clear structures and simple problem-solving characteristics (Azarmand & Neishabouri, 2009; Beaumont, 1981; Fard & Hajaghaei-Keshteli, 2018). Heuristic methods encompass a range of algorithms based on empirical rules, strategies, and evolutionary processes, including simulated annealing algorithm (Murray & Church, 1996), swarm intelligence algorithm (Gupta et al., 2017), multi-objective improved immune algorithm (Bolouri et al., 2018), ant colony optimization algorithm (Ting & Chen, 2013), among others. They all involve bio-inspired search characteristics (Kar, 2016). With the increase of the resource scale, the solution space of the spatial resource allocation optimisation problem has a high-dimensional multi-peak combination explosion, which makes the solution with the heuristic methods easy to fall into a local search (Zhang, 2022). Based on the variant hybrid innovations and computing framework updates, scholars have improved the heuristic method of facility configuration by including, e.g. differential evolution hybrid particle swarm (Hameed et al., 2020; S. Wang et al., 2022) and fast parallel transmission algorithm (Lei et al., 2016) to alleviate the impact of facility scale on computational complexity to achieve great results.

Despite such insights, all heuristic methods still follow the continuous spatial progressive search strategy based on the deterministic coding logic that mostly takes theoretical scenarios as the analysis object. This phenomenon makes it difficult to adapt to the spatial correlation (Cron & Sherman, 1962; Kelejian & Robinson, 1995) and spatial heterogeneity (De Marsily et al., 2005; Habin et al., 1998) of geographical phenomena in the urban coupling environment constraints for mixing alternations. With the continuous development of information technology and the increasing popularity of information storage devices, industries have been accumulating large amounts of data, e.g. their taxi trajectories (Al-Dohuki et al., 2016; Liu et al., 2019), taxi orders (Tong et al., 2021; Zhang et al., 2017), and Point of Interest (Liu et al.) data (Liu et al., 2013; Yuan et al., 2013). The key and open question requiring further research is how to make full use of such data and discover their rules and strategies to effectively solve the spatial resource optimisation problems because the conventional approaches, including the precise and heuristic methods, fail to provide the results with more promising insights.

In recent years, reinforcement learning (RL) methods have made breakthroughs in games (Kaiser et al., 2019; Lample & Chaplot, 2017; Littman, 1994), Go (Bouzy & Chaslot, 2006; Silver et al., 2018; Silver et al., 2007), autonomous driving (J. Chen et al., 2019; Kiran et al., 2021; Sallab et al., 2017; Shalev-Shwartz et al., 2016), robot control (Brunke et al., 2022; Johannink et al., 2019; Kober et al., 2013), pedestrian simulation (Mu et al., 2023; Xu et al., 2020) etc. It has become a new research boom in the era of artificial intelligence and has also brought new opportunities to optimise spatial resource allocation (Barto & Sutton, 1997; Feriani & Hossain, 2021). Reinforcement learning can achieve nearly real-time decision-making since its training to generate effective models can be performed offline (Levine et al., 2020; Ramstedt & Pal, 2019; Skordilis & Moghaddass, 2020). It also doesn't need to model the scene logic, and it can learn the experience of data interactions and gradually discover their rules and strategies to obtain effective models (Degris et al., 2012; Strehl et al., 2006). Moreover, when combined with deep learning methods, reinforcement learning provides more ability for large-scale data processing and discovering and extracting their low-level features providing efficient results (Arulkumaran et al., 2017; Li, 2017). In general, all such characteristics have made reinforcement learning more appropriate and robust for handling spatial resource allocation problems with big data.

This paper aims to conduct a comprehensive investigation into the application of reinforcement learning in the field of spatial resource allocation. To highlight the differences in characteristics and optimization objectives among various application scenarios, we categorize the applications of reinforcement learning in spatial resource allocation into three major classes: a) static demand resource allocation, b) static resource allocation, and c) dynamic resource allocation. In Chapter 2, we provide an introduction to the basic concepts and algorithms of reinforcement learning and deep reinforcement learning, elucidating the advantages of their application in spatial resource allocation. Chapter 3 to 5 review the latest research progress in the application of reinforcement learning within the three aforementioned categories, considering the distinct characteristics and optimization objectives of spatial resource allocation scenarios. Chapter 6 outlines some significant open issues in the application of reinforcement learning in spatial resource allocation, aiming to provide insights for future research directions. Finally, in Chapter 7, we summarize and conclude this paper. The content structure of this review is depicted in Figure 1.

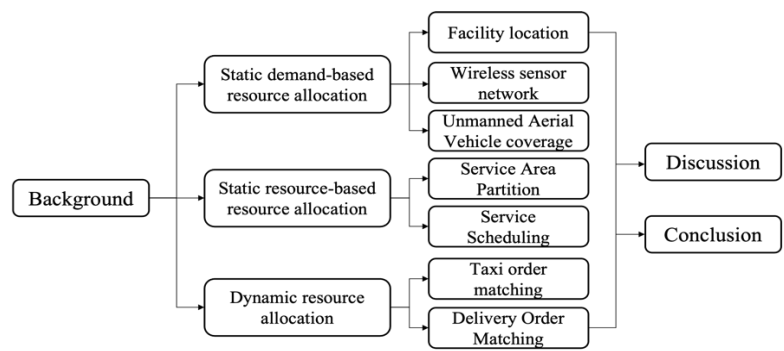

**Fig. 1** Chapter framework of this review

## 2 Background

Artificial intelligence (AI) has achieved unprecedented development with the advent of the era of big data and the continuous improvement of computer computing power (Duan et al., 2019; O'Leary, 2013). From the initial Turing test (Moor, 1976; Pinar Saygin et al., 2000) to the subsequent conception of artificial intelligence, scientists have been working hard to make computers or robots have "intelligence" and be able to learn to observe and act according to "consciousness" like humans. AI has even surpassed humans in tasks such as in the big data analysis (Kibria et al., 2018; Zhu, 2020), chess (Hassabis, 2017; Schrittwieser et al., 2020), disease diagnosis (Kumar et al., 2022; Shen et al., 2019) and video game (Perez-Liebana et al., 2016; Skinner & Walmsley, 2019). AI technology can also be widely used in other applications, such as weather forecasting (Anđelković & Bajatović, 2020; Baboo & Shereef, 2010; Liu et al., 2014), material design (Feng et al., 2021; Guo et al., 2021), recommendation system (Verma & Sharma, 2020; Zhang et al., 2021), machine perception and control (Chalmers et al., 1992; Wechsler, 2014), autonomous driving (Atakishiyev et al., 2021), face recognition (Beham & Roomi, 2013; Sharma et al., 2020), speech recognition (Al Smadi et al., 2015; Nassif et al., 2019), and dialogue systems (Deriu et al., 2021; Ni et al., 2023). An AI system needs to have the ability to learn from raw data, which Arthur Samuel (Samuel, 1959) calls Machine Learning. The usual process for AI to solve problems is to design targeted pattern recognition algorithms to extract valid features from raw data and then use these features with machine learning algorithms. Machine learning can be divided into supervised learning (Nasteski, 2017), unsupervised learning (Alloghani et al., 2020), and reinforcement learning (Sutton, 1992). Among the key differences between reinforcement learning and others is that reinforcement learning is a self-supervised learning method (Xin et al., 2020). On one hand, the agent undergoes training based on action and reward data, optimizing its action strategies. On the other hand, it autonomously interacts with the environment, receiving feedback based on the outcomes of state transition. Presently, reinforcement learning has demonstrated exceptional performance across various domains including robot control (Brunke et al., 2022), path planning (Panov et al., 2018), video game (Jaderberg et al., 2019), autonomous driving (Kiran et al., 2021), and more.

## 2.1 Reinforcement learning
### 2.1.1 Algorithm composition

The main body of reinforcement learning has two parts, the agent and the environment. It does not require supervised signals to learn but relies on the agent's feedback reward signal in the environment. The state and actions of the agent are corrected according to the feedback signal so that the agent can gradually maximise the reward. Finally, reinforcement learning can have a strong self-learning ability. A standard reinforcement learning algorithm consists of four elements: policy function, reward function, value function, and an environment model:

**Policy function:** This function defines the behaviour of the learning agent at a specific time, also known as the mapping from environmental states to actions. The probability distribution function or probability density function maps the environmental state set $S$ to the behaviour set $A$, which guides the agent in choosing the best action.

**Value function:** The value function is the expected return of states, which predicts future rewards. Reinforcement learning uses it to evaluate the quality of the state. This predicted value is closely related to the agent's policy, so the value function refers to the value function under a certain policy.

**Reward Function:** The reward function is the evaluation standard of the agent, and defines the goal in the reinforcement learning problem. The agent should try different actions to obtain high rewards as much as possible. It will generate an immediate reward $R_t$ based on the environmental state $S_t$, the action made by the agent at each time step, and send it to the agent.

**Environment model:** It is different from the real environment, it is a simulation of the external environment and is responsible for perceiving changes in the environment. It allows inferring the behaviour of the external environment, that is, what kind of feedback an agent might get for a certain action in a certain state. The environment model can predict this feedback, while the actual feedback is given by the environment based on state and action. So the closer the environment model is to the environment, the more accurate it will be. The environment model is more used for planning, that is, the agent "thinks and plans" before taking action.

**2.1.2 Algorithm framework**

In reinforcement learning, when the agent takes a particular action, it doesn't always lead to a specific state. Generally, the likelihood of transitioning to a state after an action is represented through a state transition model. The probability of the environment transitioning to the subsequent state within the actual environment process depends on multiple prior environment states. To streamline the environment's state transition model, reinforcement learning assumes that the probability of moving to the next state solely relates to the preceding state. Consequently, the entire reinforcement learning process can be simplified into a Markov Decision Process (Sutton et al.), which serves as the fundamental framework for reinforcement learning. A MDP is represented by a five-tuple $< S, A, P, R, \gamma >$, where each element signifies:

$S$ represents the set of states, encompassing all possible states that an agent can explore within the environment. $s$ denotes the current state of the agent at a given time, while $s'$ signifies the subsequent state of the agent at the next time step.

$A$ represents the set of actions, encompassing all possible actions that an agent can take based on the environmental state. $a$ denotes the action taken by the agent at the current time step.

$P$ is the state transition function, defined as follows:

$$P_{ss'}^a = P\{S_{t+1} = s' | S_t = s, A_t = a\} \tag{1}$$

$R$ represents the reward function, which signifies the expected reward obtained by the agent after taking action $A_t$ based on state $S_t$, and at time $t + 1$. The formula is expressed as follows:

$$R_s^a = E(R_{t+1} | S_t = s, A_t = a) \tag{2}$$

$\gamma$ represents the discount factor, which is the proportion of the value of future rewards at the current moment.

Reinforcement Learning operates as a MDP where an agent learns to make decisions aimed at achieving specific goals through interactions with its environment. In this process, the agent observes environmental states, chooses actions guided by a particular strategy, and receives corresponding rewards or penalties from the environment. The agent's goal is to maximize long-term rewards by experimenting with diverse strategies. The following Figure 2 outlines the core process of reinforcement learning.

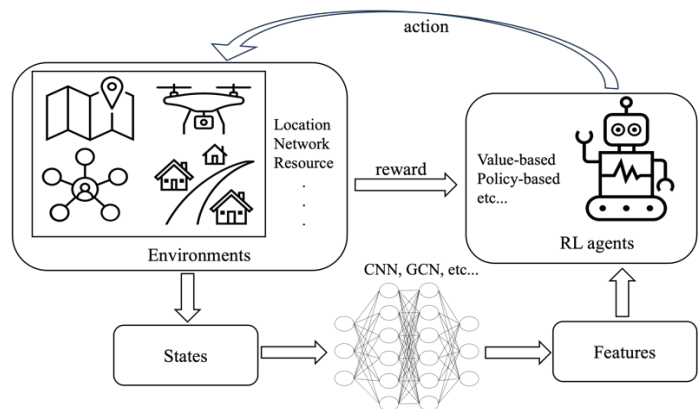

Fig. 2 Overall reinforcement learning framework

## 2.2 Category

Based on current research, reinforcement learning algorithms are categorized based on diverse criteria such as function-based approaches, and the number of agents in the environment. Function-based classification groups reinforcement learning algorithms into three categories: value-based, policy-based, and Actor-Critic methods used to update learning policies. Concerning the number of agents in the environment, reinforcement learning algorithms are classified into single-agent and multi-agent categories. In a multi-agent system where multiple agents interact with the environment, each agent still pursues its reinforcement learning objective. The alteration in the environment's overall state relates to the collective actions of all agents. Therefore, considering the impact of joint actions becomes crucial in the process of agent policy learning.

### 2.2.1 Value-based RL

The value-based reinforcement learning algorithm implicitly shapes the optimal policy by deriving the optimal value function and selecting actions corresponding to maximum value functions. Notable algorithms in this category include Q-learning (Watkins & Dayan, 1992), SARSA (Zhao et al., 2016), and Deep Q-Network (DQN) (Fan et al., 2020). DQN merges the Q-learning algorithm with a deep neural network, usually employing DNN or CNN to build a model and the Q-learning algorithm for training. This method effectively addresses the computational inefficiency and limited data memory concerns of Q-learning. However, due to the overestimation drawbacks in DQN, various optimization algorithms emerged, such as Deep Double Q-learning Network (DDQN) (Van Hasselt et al., 2016), Dueling DQN (Wang et al., 2016), DQN algorithms with dynamic frame skipping (Srinivas et al., 2016), Prioritized Experience Replay (PER) (Schaul et al., 2015), Noisy DQN (Fortunato et al., 2017), Distributional DQN (Dabney et al., 2018), Rainbow DQN (Hessel et al., 2018), etc. Although value-based algorithms offer benefits like high sample efficiency, low variance in value function estimates, and resilience to local optima, they commonly struggle with continuous action space problems. Figure 3 illustrates the process of value-based reinforcement learning:

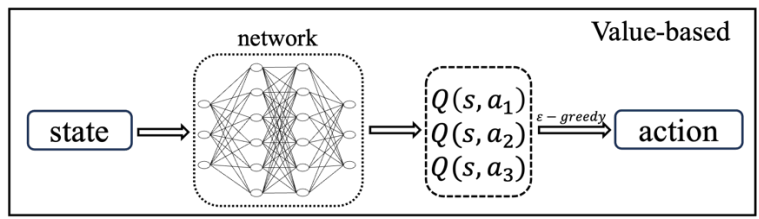

**Fig. 3** Basic framework of value-based reinforcement learning algorithms

### 2.2.2 Policy-based RL

Given the limitations of reinforcement learning methods based on the value function concerning continuous action space parameters and stochastic policy issues, researchers have proposed various policy-based reinforcement learning approaches. In policy-based algorithms, the agent directly produces the probability of potential actions for the subsequent time step and selects actions based on these probabilities. These algorithms parameterize the policy, utilizing the expected cumulative return as the objective function, and optimize this function through gradient policy methods (Silver et al., 2014). The stochastic policy search method learns the parameterized policy directly based on policy gradients. It bypasses the need to solve the action space value maximization optimization problem, making it more suitable for addressing high-dimensional or continuous action space problems. Notable algorithms in this category include REINFORCE (Williams, 1992), Trust Region Policy Optimization (TRPO) (Schulman et al., 2015), Proximal Policy Optimization (PPO) (Schulman et al., 2017), Distributed PPO (Zhang et al., 2019), Trust-PCL (Nachum et al., 2017), etc. In contrast to random strategies, deterministic strategies determine an action uniquely for a specific state. Representative algorithms encompass Deterministic Policy Gradient (DPG) (Srinivas et al., 2016), Deep Deterministic Policy Gradient (DDPG) (Lillicrap et al., 2015), TD3 (Lillicrap et al., 2015), and so on. However, policy-based reinforcement learning exhibits certain drawbacks: it can be computationally intensive and entails extended iteration times in addressing complex problems. Figure 4 illustrates the process of policy-based reinforcement learning:

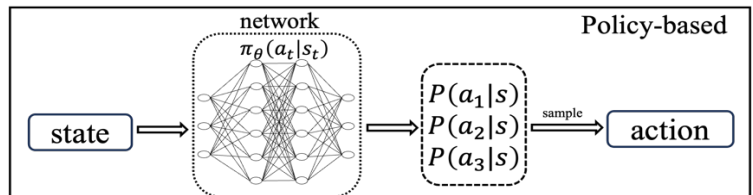

Fig. 4 Basic framework of policy-based reinforcement learning algorithms

### 2.2.3 Actor-Critic RL

The Actor-Critic algorithm combines aspects of both value-based and policy-based reinforcement learning methods. Its architecture involves two key components: the Actor, which employs policy methods to approximate the policy model by generating actions and interacting with the environment, and the Critic, which employs value methods to assess the advantages and disadvantages of actions and approximates the value function. Subsequently, the Actor optimizes the action probability function based on the Critic's evaluations, guiding the agent to choose optimal actions. This approach conducts policy evaluation and optimization using the value function while refining the policy function to enhance the accuracy of state value representation. These intertwined processes converge to derive the optimal policy. In recent years, several Actor-Critic algorithms have emerged, including Advantage Actor-Critic (A2C) (Grondman et al., 2012), Asynchronous Advantage Actor-Critic (A3C) (Mnih et al., 2016), and Soft Actor-Critic (Rosaci & Sarnè) (Haarnoja et al., 2018), etc. However, adjusting the parameters of the Actor-Critic algorithm can be challenging, and the consistency of results when applied may not be guaranteed. Figure 5 illustrates the process of Actor-Critic reinforcement learning：

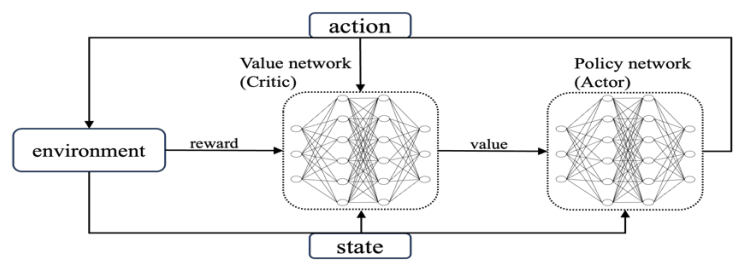

Fig. 5 Basic framework of Actor-Critic reinforcement learning algorithms

### 2.2.4 Multi-agent RL

Complex real-world scenarios often demand collaboration, communication, and confrontation among multiple agents, such as production robots (Giordani et al., 2013), urban traffic lights (Wu et al., 2020), E-commerce platform (Rosaci & Sarnè, 2014), all of which constitute typical multi-agent systems.The basic idea of multi-agent reinforcement learning is the same as that of single-agent reinforcement learning. It is to let the agent interact with the environment and then learn to improve its strategy according to the reward value obtained to obtain the situation in the environment. However, in contrast to single-agent environments, multi-agent settings are notably more intricate. The state space and the connectivity of action space in these environments grow exponentially with the number of agents involved. Each agent must navigate this dynamic environment, involving interactions with other agents, thereby intensifying the learning complexity. Additionally, as multiple agents engage with the environment concurrently during the learning phase, the actions of one agent can instigate alterations within the environment, thereby influencing its own decision updates. Furthermore, the influence of other agents on the environment perpetually impacts subsequent decisions. The evolution of multi-agent reinforcement learning is deeply intertwined with game theory, considering its resemblance to a stochastic or Markov game in the learning process.

Classic multi-agent reinforcement learning algorithms based on game theory include Nash Q-Learning (Hu & Wellman, 2003),Team Q-learning (Cassano et al., 2019), Minimax-Q-Learning (Zhu & Zhao, 2020), Friend-or-Foe Q-learning (FFQ) (Littman, 2001), etc. The development of deep learning has broken through the characteristics of small-scale and simple problems applicable to classical multi-agent reinforcement learning algorithms. Through the extension or improvement of the single-agent reinforcement learning algorithm, the researchers propose Independent Double Deep Q-Network (IDDQN) (M. Wang et al., 2022), Mean Field Multi-Agent Reinforcement Learning (MFMARL) (Yang et al., 2018), and Deterministic Policy Gradient (MADDPG) (Lowe et al., 2017), etc. The advantage of multi-agent reinforcement learning lies in handling complex interactive environments, yet it faces challenges related to high-dimensional state spaces and instability.The process of multi-agent reinforcement learning is shown in Figure 6:

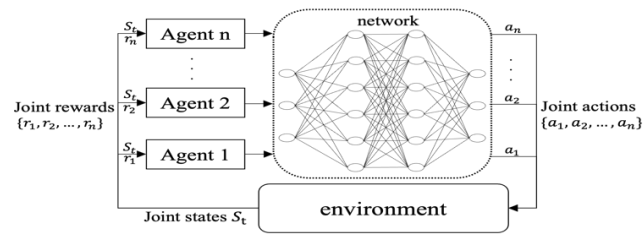

**Fig. 6** Basic framework of multi-agent reinforcement learning algorithms

## 2.3 Potentiality of applying reinforcement learning to spatial resource allocation

The problem of spatial resource allocation is that planners allocate spatial resources reasonably and effectively according to the current spatial allocation of resources, the set constraints and the goals to be achieved. Scientific and rational allocation of spatial resources can effectively improve the utilisation efficiency of resources, reduce the operating cost of facilities, and improve the efficiency of output. The problem of spatial resource allocation is a typical NP-Hard problem, considering multiple challenges such as complex problem specification description and interleaved spatial structure. The potential application of reinforcement learning in spatial resource allocation stems from its fundamental principle of learning through the interaction between an agent and its environment. In resource allocation problems, the agent (representing decision-makers or systems) needs to make a series of decisions based on the environmental state to achieve optimal resource utilization.

In the context of spatial resource allocation problems, the state may encompass information regarding the location, quantity, availability of resources, and environmental attributes related to resource allocation. Actions refer to the decisions made by an agent concerning resource allocation, such as determining which areas or locations resources should be allocated to. Rewards typically assess the agent's decision-making based on resource utilization efficiency, coverage scope, cost savings, or other specific objectives. Through continual interaction and experimentation with the environment, the agent adjusts its behavioral strategies based on received reward signals, progressively learning the optimal resource allocation scheme. This learning process involves striking a balance between exploration and exploitation; the agent needs to explore novel decision choices to discover better solutions while leveraging existing knowledge to enhance resource utilization efficiency.

Therefore, based on the principles of reinforcement learning, employing this technique to address various spatial resource allocation problems becomes feasible. Reinforcement learning agents optimize resource allocation strategies incrementally through interactions with the environment, adapting to evolving environmental conditions and demands for more effective resource utilization.

## 3 Reinforcement learning in Static demand-based resource allocation

In this scenario, the state of demand points remains fixed and static, while the location of resource points is selectable, representing a dynamic allocation focused on resource points. Facility location and service coverage exemplify typical instances of static demand-based resource allocation scenarios, aiming to optimize the spatial and temporal distribution of resource demand and consumption within a limited space. The goal is to ensure consistency in meeting resource demands and utilization across time and space, maximizing resource service provision to a larger set of demand points whenever feasible.

### 3.1 Facility location

For facility location selection, both value-based and policy-based reinforcement learning algorithms are used. Aidan et al. (2021) and von Wahl et al. (2022) used the DQN algorithm to optimise the location of charging stations in cities. Leonie et al. used the charging stations' location, charging demand, and facility cost as state functions and set the action space of the agent as create, increase, and relocate. The reward of agent is related to the construction cost of the charging bin and output income. Aidan et al. used a supervised gradient to improve the model's accuracy, describing the state as a 2D grid model containing the predicted demand value. They used placing a charger in one grid cell as the action space of agent, sum of the expected demand of a hypothetical charger in a grid cell, and the area covered by the charger as the reward factor. Liu et al. (2023) proposed the policy-based PPO-Attention model, which integrates an attention mechanism into the PPO algorithm, enhancing the algorithm's ability to recognize and comprehend the intricate interdependencies among different nodes within the network. Zhao et al. (2023) proposed a recurrent neural network (RNN) with an attention mechanism to learn model parameters and determine the optimal strategy in a completely unsupervised manner.

In optimizing the coverage range of Unmanned Aerial Vehicle (UAV) base stations to meet diverse ground user needs, Gopi et al. (2021) proposed a reinforcement learning algorithm based on Q-learning to guide the selection of UAV base stations. This study restricts base station movement to grid points in a square grid, reducing the number of states and allowing drone base stations to move in fixed incremental distances to maintain a consistent distance between stations. Bhandarkar et al. (2022) utilized DQN algorithm to select UAV base station locations more reasonably. They used the UAV's position and covered users as states, while the number of newly covered users, user coverage, and whether the UAV exceeded boundaries were treated as rewards. By comparing reward-based greedy algorithm with DQN algorithm, this study concludes that DQN algorithm outperforms greedy algorithm in both coverage and latency performance.

## 3.2 Wireless sensor network

When exploring the application of reinforcement learning in resource coverage optimization within wireless sensor networks, Nie et al. (1999) and El-Alfy et al. (2006) utilized real-time Q-learning combined with neural network representations to address dynamic wireless resource allocation problems. Concurrently, Seah et al. (2007) and Renaud et al. (2006) employed distributed reinforcement learning algorithms such as IL, DVF, COORD, optimistic DRL, and FMQ to optimize the resource coverage and energy consumption in wireless sensor networks. Additionally, Tamba et al. (2021) utilized unmanned aerial vehicles as sensing components within a mobile wireless sensor network, establishing a three-dimensional grid environment and deploying the wireless sensor network as a whole to perform the resource coverage task within a target area.

Addressing coverage gaps in wireless sensor networks, researchers have also employed reinforcement learning for coverage hole repair. Hajjej et al. (Hajjej et al., 2020) proposed a distributed reinforcement learning approach where each intelligent agent selected a combination of node relocation and perception range adjustment actions. Simulation results demonstrate that the proposed method can sustain overall network coverage in the presence of random damage events. Furthermore, Philco et al. (Philco et al., 2021) initially employed a multi-objective black widow optimization algorithm and Tsallis entropy-enabled Bayesian probability (TE2BP) algorithm for dynamic sensor node scheduling. Subsequently, they utilized a multi-agent SARSA to determine the optimal mobile node for repairing coverage holes.

## 3.3 Unmanned Aerial Vehicle coverage

The coverage problem for UAVs doesn't necessitate serving all demands but rather expanding the coverage as much as possible within a certain number of service facilities. Liu et al. (2018) proposed DRL-EC3 based on DDPG, considering communication coverage, fairness, energy consumption, and connectivity, guided by two deep neural networks (DNN). Wang et al. (2019) utilized the DDQN algorithm to determine UVA positions. The model used UVA positions as the state function, with action spaces comprising five directional actions, and reward functions based on coverage and capacity ratios, achieving optimal real-time capacity by altering UVA allocation positions with moving ground terminals. Xiao et al. (2020) proposed a distributed dynamic area coverage algorithm based on reinforcement learning and γ information graph. The γ information graph can transform the continuous dynamic coverage process into a discrete γ point traversal process while ensuring no hole coverage. When agent communication covers the entire target area, agent can obtain the global optimal coverage strategy by learning the entire dynamic coverage process. When the communication does not cover the entire target area, the agent can obtain a locally optimal coverage strategy. Bromo et al. (2023) presented PPO method for UAV fleet coverage planning in unexplored areas with obstacles, aiming to reduce steps and energy needed for full coverage while avoiding collisions.

There are also researchers using multi-agent reinforcement learning to solve this problem, Pham et al. (2018) addressed the UVA service environment using a 3D grid, defining UAV state as its approximate location in the environment. They applied the multi-agent Q-learning algorithm to comprehensively cover service resources in unknown areas of interest while minimizing overlapping fields of view. Meng et al. (2021) introduced the MADDPG algorithm, treating multiple UVAs as agents, considering UVA service coverage rate and location as states, and setting travel distances and directions as agent actions, ensuring good dynamic coverage while maintaining agent group connectivity.

## 3.4 Chapter summary

In scenarios involving static demand points for resource allocation, due to the fixed positions of these points, reinforcement learning often uses discrete location information or grid data as states. The agent's action selection is abstracted into the geographical positions where resources can be placed. This approach simplifies the complex real-world environment for the model to comprehend and extract feature information more easily. However, this method might overlook topological information present in reality, such as natural barriers or political boundaries. Consequently, there can be significant errors in measuring distances and accessibility between nodes. Relevant papers tackling these tasks are summarized in Table 1.

**Table 1** Summary of RL applications to static demand-based resource allocation

| Application | Reference | Algorithm | State | Action | Objective |
|---|---|---|---|---|---|
| Facility location | Aidan et al. (2021) | DQN | 2D grid model contains demand | Place charger in grid | Sum of expected demand |
| | Von Wahl et al. (2022) | DQN | Location, demand, costs | Create, increase, relocate | Benefit, cost |
| | Liu et al. (2023) | PPO-attention | Location, demand, costs | Create increase relocate | Profit cost fairness |
| | Zhao et al. (2023) | RNN-attention | Number of installed chargers, completed time steps | Charging station location | Quality of service |
| | Gopi et al. (2021) | Q-learning | 2D grid model with locations | Five directions | Data transfer rate |
| | Bhandarkar et al. (2022) | DQN | UVA location, Users covered | Eight directions | Covers number of new users, user coverage status |
| Wireless sensor network | Nie et al. (1999) | Q-learning | Available channels | Assign channel | Cost |
| | El-Alfy et al. (2006) | Q-learning | Available channels | Reject, admit | Cost |
| | Seah et al. (2007) | IL, DVF, COORD | 2D grid with three statuses | Hibernate, sense | State transition, bonus gain |
| | Renaud et al. (2006) | Q-learning, DVF, optimistic DRL, FMQ | Each grid's sensing status | Different modes | Coverage, energy consumption |
| | Tamba et al. (2021) | Q-learning | 3D grid contains position | Six orientations | Coverage, overlap |
| | Hajjej et al. (2020) | Distributed payoff-based Q-learning | G (V, E) with attributes | Possible position | Coverage |
| | Philco et al. (2021) | Muti-agent SARSA | Distance, Node lifetime, Coverage level | Position, range | Coverage |
| Unmanned Aerial Vehicle coverage | Liu et al. (2018) | DRL-EC3 | Coverage state, energy consumption | Directions, distance | Coverage score, fairness, energy consumption |
| | Wang et al. (2019) | DDQN | UVA location | Five orientations | Coverage, capacity ratios |
| | Xiao et al. (2020) | Q-Traversal | γ-information map | Adjacent position | Position, information value |
| | Bromo et al. (2023) | PPO | Agent's selection, obstacle | Four directions | Coverage |

| | | | | |
|---|---|---|---|---|
| Pham et al. (2018) | Equilibrium-based Q-learning | 3D grid contains position | Six orientations | Coverage |
| Meng et al. (2021) | MADDPG | UVA location, Coverage | Orientation, distance | Coverage, connectivity penalty |

## 4 Reinforcement learning in Static resource-based resource allocation

In this scenario, the state of resource points remains static, while demand points are subject to dynamic changes. The characteristics or quantity of resources remain constant throughout the allocation process, yet the positions or quantities of demand may fluctuate with changing demands. Dividing service spaces and scheduling services represent instances of static resource-based allocation scenarios, where rational partitioning or arrangement of demand areas enables resources to maximize their service efficacy within a confined space.

### 4.1 Service Area Partition

Klar attempted to solve factory layout planning problems using reinforcement learning, proposing different state spaces and reward setups for various optimization objectives. Initially, they used the DDQN to address a layout scenario involving four functional units, optimizing for transportation time (Klar et al., 2021). Then, for scenarios with numerous functional units, they introduced a novel state representation method combined with action masking, optimizing the action selection process to ensure scalability and reduce training time (Klar, Hussong, et al., 2022). Researchers later focused the state space more on specific details and additional information for placing the next functional unit when optimizing for material flow and energy consumption (Klar, Langlotz, et al., 2022). Additionally, Klar proposed a comprehensive framework for reinforcement learning-based factory layout planning, integrating graph neural networks (GNN) (Scarselli et al., 2008) with DDQN to enhance feature extraction for states, applicable in both initial factory layout planning and restructuring phases (Klar et al., 2023). Other researchers, including Wang et al. (2020), Di et al. (2021), and Ribino et al. (2023), utilized reinforcement learning algorithms like PEARL, MCTS, and MORL to propose improved furniture placement solutions in households. In particular, Wang et al. transformed a 3D internal graphics scene into two 2D simulation scenes, establishing a simulation environment where two reinforcement learning agents cooperatively learned the optimal 3D layout through a MDP formulation. Kim et al. (2020) addressed the shipyard layout problem to minimize the use of transport aircraft during rearrangements using the A3C algorithm. This research involved two agents in the decision-making process: the transporting agent and the locating agent, each with distinct states, actions, and reward functions. The transporting agent considered stockyard blocks' location and remaining time, choosing the next block arrangement, with the reward function based on the number of blocks moved. Meanwhile, the locating agent dealt with blocks and transporters in the stockyard, deciding whether to move a transporter or carry a block, with the reward based on the success or failure of block export.

### 4.2 Service Scheduling

The dynamic charging scheduling for electric vehicles aligns with static resource allocation challenges. These strategies select charging stations for vehicles on the move, aiming to cut overall charging times while easing grid pressure. Researchers have used reinforcement learning to propose solutions, varying in state designs, set constraints, and optimization objectives. States typically include Battery State of Charge (SOC) (Aylor et al., 1992), charging demands, and costs for action selection, crucial for managing distributed energy systems. Optimization often centers on energy and charging costs, attempting to minimize imbalances between generation and consumption. In terms of constraints, the majority of proposed methods utilize energy network constraints to ensure that the objectives of reinforcement learning agents are not the ultimate optimal states but rather the best states practically attainable within the existing electric vehicle charging network. Some studies also define battery parameters such as capacity and charging rates as constraints to provide a genuine representation of battery behavior within the model. Q-learning and its deep extension, DQN, emerge as effective solutions, and the combination of DQN with DDPG can address the challenges posed by high dimensionality and discretization. Additionally, multi-agent reinforcement learning solutions have been proposed for dynamic electric vehicle charging scheduling problems, although their computational costs are high, and convergence poses significant challenges, limiting their practical application. Due to the extensive research methodologies in this field of reinforcement learning, several related review articles are recommended for further exploration (Abdullah et al., 2021; Fescioglu-Unver & Aktaş, 2023; Qiu et al., 2023; Shahriar et al., 2020).

### 4.3 Chapter summary

When dealing with static resource allocation scenarios, reinforcement learning demonstrates certain advantages. As the resource locations remain unchanged, it reduces the impact of environmental dynamics on the system, making the model more stable and predictable. It can optimize resource distribution to meet long-term demands and better plan for their prolonged utilization. However, the fixed nature of resources in the scenario may pose challenges, making it difficult for the learning model to adapt to new demands or environmental changes, requiring more flexible decision strategies to cope with such variations. Relevant papers tackling these tasks are summarized in Table 2.

**Table 2** Summary of RL applications to static resource-based resource allocation

| Application | Reference | Algorithm | State | Action | Objective |
|---|---|---|---|---|---|
| Service Area Partition | Klar et al. (2021) | DDQN | 2D grid with status | Location and corner | Transportation time |
| | Klar et al. (2022) | DDQN | Point with position status, occupancy | Location and corner | Validity of action |
| | Klar et al. (2022) | DQN | Status information of next functional unit | Location and corner | Material flow, energy consumption |
| | Klar et al. (2023) | GNN+DDQN | Layout with flow characteristics | Placement options | RTT, RUFU, RUMF, RTC, RTI, RMS, RFBC, RCMF, RS |
| | Wang et al. (2020) | PEARL | Scene layout | Four directions | Distance, Arrival, Leave, Success |
| | Wang et al. (2021) | MCTS | 3D grid contains position | Four directions | Ground truth position |
| | Di et al. (2021) | DQN | 2D grid layout | Object, Four directions, obstacle | Target layout |
| | Ribino et al. (2023) | MORL | Arrangement of furniture set | Four directions, Rotate, No Move | Indoor environmental quality, distance |
| | Kim et al. (2020) | A3C | Blocks in stockyard with status | Select block | Number of blocks moved during rearrangement stage |
| Service Scheduling | Shahriar et al. (2020) | Review | | | |
| | Abdullah et al. (2021) | Review | | | |

| | | |
|---|---|---|
| Fescioglu-Unver et al. (2023) | Review | |
| Qiu et al. (2023) | Review | |

## 5 Reinforcement learning in Dynamic resource allocation

In this scenario, both resource points and demand points exhibit dynamic variability, where their positions, quantities, or characteristics may change over time. For instance, considering mobile services in urban settings such as bike-sharing (DeMaio, 2009), taxi service (Yang & Wong, 1998), or mobile application-based services (Ervasti & Helaakoski, 2010), the positions and quantities of resources might fluctuate in response to user demands, which themselves can vary based on both time and location. In such cases, resource allocation necessitates real-time adjustments based on the dynamic changes in demand and resource availability.

### 5.1 Taxi order matching

In the taxi order matching problem, since both supply and demand are dynamic, the uncertainty comes from the constant location of the demand point, initial locations of the driver and required travel times. In this resource allocation scenario, the highly dynamic supply and demand position relationship and the acquisition of long-term revenue are complex challenges. Reinforcement learning optimised for this scenario aims to improve the service quality of the resource matching system and the total income of drivers over a long period of time. When researchers use reinforcement learning to build a taxi order matching model, they can be divided into two modes: single-agent training and multi-agent training according to the settings of agents.

1) Single-agent model

In single-agent model, all agents are defined with the same state, action space, and reward definition. The researchers trained the agent with the experience trajectories of all drivers and applied them to generate matching policies. In this model, although the system is multi-agent from a global perspective, only a single agent is considered in the training phase. The most commonly used state elements include current vehicle location, passenger location, and order details (Al-Abbasi et al., 2019; Holler et al., 2019; Tang et al., 2019; Wang et al., 2018; Xu et al., 2018). These order details include, in addition to existing orders, forecasted demand information derived from forecasting models. For example, Zhou et al. (2023) proposed the ATD3-RO algorithm, which combines adaptive Twin Delayed Deep Deterministic Policy Gradient with robust optimization to perform order prediction in uncertain passenger scenarios. Yang et al. (2021) modelled each demand as an agent and trained a value network to estimate the demand rather than the worker's value, further performing a separate many-to-many matching process based on the learned value.This approach aims to facilitate taxi order matching based on future order predictions even in situations with passenger uncertainties. Additionally, some methods involve environment discretization into grids and utilize graph-based approaches to depict state characteristics (Gao et al., 2018; Haliem et al., 2021; Rong et al., 2016; Verma et al., 2017). In more specific scenarios, such as operations involving new energy taxis, researchers have included the battery storage of vehicles as part of the state (Shi et al., 2019; Tu et al., 2023).

Trip price and profit have become the ultimate optimization objectives in this scenario. However, in setting rewards, researchers consider various factors, such as travel distance, waiting time, and the probability of successful transactions. Some improved algorithms based on Q-earning have been employed in this model. Wang et al. (2019) introduced the dynamic bipartite graph matching (DBGM) problem, taking a holistic system perspective. They trained an agent that encapsulated the entire request list and employed a restricted Q-learning algorithm to optimize decision duration, resulting in near-maximized rewards. Sanket Shah et al. (2020) proposed a Neural Network-based Approximate Dynamic Programming (ADP) framework for a carpooling system, where ADP and neural networks were employed to learn approximate value functions, and DQN was utilized to stabilize the neural ADP. Guo et al. (2020) incorporated vehicle scheduling and route planning, using DDQN to balance passenger service quality against historical data-derived system operating costs. Specifically, the system considers reassigning idle vehicles to optimize vehicle route decisions. Then, through dynamic programming, a vehicle allocation plan is suggested based on learned values from vehicle routing. Wang et al. (2023) addressed passenger transfer issues in ride-sharing scenarios, allowing passengers to transfer between vehicles at transfer stations. By combining DQN with Integer Linear Programming (Verma & Sharma) (Schrijver, 1998), they employed ILP to achieve optimal online scheduling and matching strategies for each decision stage, using DQN to learn approximate state values for each vehicle. This combination introduced specific policies to limit state space and reduce computational complexity. Tu et al. (2023) utilized a spatiotemporal NN approach to extract taxi demand patterns, combining this with DDQN to form a Spatiotemporal Double Deep Q Network (ST-DDQN) aiming to maximize daily profits.

The single-agent model simplifies decision-making and reduces computational complexity in taxi order matching, yet it may face limitations due to information constraints, insufficient collaboration, and poor adaptability to environmental changes. While it offers simplified management and computations, it cannot fully leverage information from other taxis, potentially hindering optimal overall efficiency. During significant environmental shifts, such as peak hours or unusual events leading to a surge in orders or changing traffic conditions, the single-agent model might struggle to adapt effectively. Relevant papers tackling these tasks are summarized in Table 3.

**Table 3** Summary of RL applications to taxi order matching based on single agent

| Application | Reference | Algorithm | State | Action | Objective |
| --- | --- | --- | --- | --- | --- |
| Taxi order matching (single) | Wang et al. (2018) | DQN | Location, time | Assignment to specific order, idle | Trip price |
| | Xu et al. (2018) | Tabular TD | Location, time | Assign to specific order, idle | Trip price |
| | Al-Alabbasi et al. (2019) | DQN | Location, available vehicles, demand | Dispatched, not dispatched | Number of customers, time cost |
| | Holler et al. (2019) | DQN | Matching order, repositioning driver | Order dispatch, reposition | Trip price, reposition cost |
| | Tang et al. (2019) | CVNet+transfer | Location, time static features | Options | Trip price |
| | Yang et al. | DQN+TD | Location, time, contextual features within demand | Dispatch, idle | Time cost |
| | Zhou et al. (2023) | ATD3-RO | Location, time, available vehicles, demand | Assignment to specific order, idle | Cost |
| | Rong et al. (2016) | Dynamic programming | Grid with time, direction | Move, stay | Taxi fare |
| | Verma et al. (2017) | Q-learning+MC | Grid with time-interval | Move to chosen grid | Taxi fare, traveling distance cost, time cost |
| | Gao et al. (2017) | Q-learning+TD | Grid with operating status | Move, stay, wait | Ratio of occupied mileage to |

| | | | | previous empty mileage |
|---|---|---|---|---|
| Haliem et al. (2018) | DQN | Grid with available vehicles, demand | Move to chosen grid | Number of customers served, time cost, profit, vehicle utilization |
| Shi et al. (2018) | Decentralized DQN | Location, time, remaining battery | Pass, charge, assign | Incentives, costs |
| Tu et al. (2018) | ST-DDQN | Location, time, remaining battery | Serve, charge, cruise, wait | Benefit, cost |
| Wang et al. (2019) | Restricted Q-learning | Bipartite graph | Match, not match | Sum of weights of matched pairs |
| Shah et al. (2020) | DQN+ADP | Location, time, demand | Group of users | Return |
| Guo et al. (2020) | DDQN | Position, available seats, passengers | Pick-up, drop off | Passenger QoS, cost |
| Wang et al. (2023) | DQN+ILP | Location, time | Pass, assign, reposition | Incentives, costs |

2) Multi-agent model

Since it is difficult to simulate complex interactions between drivers and orders in a single-agent setting, many researchers have also applied multi-agent reinforcement learning (MARL) in such spatial resource order allocation scenarios. The research on MARL for order matching can be categorized into three major classes: based on global feature models, multi-scale models, and model integration.

In methods based on global feature models, researchers focus on capturing overall characteristics and dynamic changes. Li et al. (2019) applied multi-agent reinforcement learning to address distributed features in point-to-point carpooling, capturing global dynamic changes in supply and demand using mean field theory. Zhou et al. (2019) and Zhang et al. (2020) utilized methods like Kullback-Leibler divergence optimization to expedite DDQN learning process and balance the relationship between vehicle supply and order demand.

The multi-scale model methods emphasize multi-layered, multi-scale decision-making processes. Lin et al. (2018) proposed a contextual multi-agent reinforcement learning framework with geographic and collaborative environments, encompassing contextual DQN and contextual multi-agent actor-critic algorithms. This framework enables coordination among numerous agents across diverse contexts. Jin et al. (2019) consider spatial grid units as working agents, grouping sets of these units into managerial agents and employing a hierarchical approach to decision-making. They utilize a multi-head attention mechanism to integrate the influence of neighboring agents and capture crucial agents at each scale.

Model integration methods primarily focus on integrating multiple models or techniques to achieve a more comprehensive and effective decision-making process. This approach aims to leverage the strengths of different models to address complex problems and enhance system performance. Ke et al. (2020) established a two-stage framework involving combinatorial optimization and multi-agent deep reinforcement learning. They dynamically determine the delay time for each passenger request using multi-agent reinforcement learning, while employing combinatorial optimization for optimal binary matching between idle drivers and waiting passengers in the matching pool. Liang et al. (2021) reconstructed the online vehicle scheduling problem by leveraging the topology of heterogeneous transportation networks, using a micro-network representation based on link nodes. They integrated the order scheduling stage with the vehicle routing stage. Singh et al. (2021) developed a multi-agent ride-sharing system using travel demand statistics and deep learning models. Each vehicle in this system makes independent decisions based on its individual impact without coordination with other vehicles.

Xu et al. (2023) unified order matching and proactive vehicle repositioning into a unified MDP model, addressing challenges of extensive state spaces and driver competition.

Multi-agent reinforcement learning presents significant advantages in the context of taxi order matching. Its flexibility and adaptability allow the system to dynamically adjust vehicle allocations, making optimal decisions based on real-time traffic and order demands, consequently enhancing overall service efficiency. The collaborative decision-making capacity helps avoid duplicate service areas and facilitates information sharing, optimizing vehicle dispatch and improving system performance. However, this approach encounters challenges: computational complexity requires substantial computing resources, handling large-scale state spaces may become arduous; additionally, issues of convergence, stability within multi-agent systems, and achieving effective communication and collaboration pose challenges, potentially impacting the stability and accuracy of the learning process and decision outcomes. Relevant papers tackling these tasks are summarized in Table 4.

**Table 4** Summary of RL applications to taxi order matching based on multi agent

| Application | Reference | Algorithm | State | Action | Objective |
| --- | --- | --- | --- | --- | --- |
| Taxi order matching (multi) | Li et al. (2019) | MARL with independent Q-learning | Location, time, available status | Assignment to specific order, idle | Trip price |
| | Zhou et al. (2019) | Double DQN | Location, available vehicles, orders | Grid index | Euler distance |
| | Zhang et al. (2020) | QRewriter-DDQN | Grid with available vehicles, orders, time interval | Assign to grid | Improvement gain |
| | Lin et al. (2018) | Contextual multi-agent A2C, contextual DQN | Number of available vehicles and orders in each grid, current time | Seven directions | Averaged revenue |
| | Jin et al. (2019) | Hierarchical MARL | Location, available vehicles, orders | Worker: ranking for match and reposition; Manager: abstract goal for workers | Income, order response rate |
| | Ke et al. (2020) | Delayed-M-DQN, Delayed-M-A2C, Delayed-M-PPO, Delayed-M-ACER | Spatio-temporal patterns of supply-demand | Delayed, not delayed | Profit, time cost |
| | Liang et al. (2021) | TDCP | Time and node of vacant vehicle | Matching a customer, routing to a node | Profit |
| | Singh et al. (2021) | MHRS | Location, time, supply-demand | Order dispatch, reposition | Gap for supply-demand, time cost |
| | Xu et al. (2023) | MAMR | Grid with time interval, supply-demand | Assign to grid | Profit |

## 5.2 Delivery Order Matching

Reinforcement learning extends beyond spatial resource matching to courier order matching. To tackle the daily large-scale matching tasks in courier management, Li et al. (2019) initially segmented the city into independent areas where each area had a fixed number of couriers collaborating on package delivery and service requests. They introduced a soft-label clustering algorithm named "Balanced Delivery Service Burden" (BDSB) to distribute packages among couriers within each area. Addressing real-time pickup requests, they proposed a model called "Contextual Cooperative Reinforcement Learning" (CCRL) to guide each courier's short-term delivery and service locations. CCRL was modeled as a multi-agent system emphasizing courier cooperation while considering the system environment. In subsequent research, they introduced the CMARL algorithm aiming to maximize the total completed pickup tasks by all couriers over an extended period (Li et al., 2020). Chen et al. (2019) introduced a framework that utilized multi-layered spatio-temporal maps to capture real-time representations within service areas. They modeled different couriers as multiple agents and used PPO to train corresponding policies, enabling system agents to assign tasks in mobile crowdsourcing scenarios. Zou et al. (2022) presented an Online To Offline (O2O) order scheduler based on DDQN, intelligently assigning orders to couriers based on new order status and the collective courier status. Jahanshahi et al. (2022) explored food delivery services using different variants of DQN for a specific group of couriers to meet dynamic customer demands within a day, paying particular attention to the impact of limited available resources on food delivery. The results offered valuable insights into the courier allocation process concerning order frequencies on specific dates.

### 5.3 Chapter summary

In dynamic resource allocation, the choice between single-agent or multi-agent reinforcement learning depends on the complexity of the actual scenario, the variability of demands, and the requirements for system performance and stability. Single-agent systems are suitable for relatively simple and stable environments, while multi-agent systems are more suitable for complex and dynamic scenarios, offering better utilization of resources through enhanced collaboration. Relevant papers tackling these tasks are summarized in Table 5.

Table 5 Summary of RL applications to delivery order matching

| Application | Reference | Algorithm | State | Action | Objective |
| --- | --- | --- | --- | --- | --- |
| Delivery Order Matching | Li et al. (2019) | CCRL | Location, time, supply-demand | Assign to grid | Number of completed tasks |
| | Li et al. (2020) | CMARL | Location, time, supply-demand | Assign to grid | Number of completed tasks |
| | Chen et al. (2019) | PPO | Location, time, supply-demand | Assign to grid | Total price |
| | Zou et al. (2022) | Double-DQN | Location, time, supply-demand | Dispatch to courier | Orders' completion time |
| | Jahanshahi et al. (2022) | DQN | Location, time, supply-demand | Assign, reject order, move towards restaurant, depot | Delivery time |

## 6 Discussion

While researchers have extensively utilized reinforcement learning to devise algorithms and model frameworks for resolving spatial resource allocation problems across various real-world scenarios, the construction of a more practical framework continues to pose challenges. We aim to briefly outline the primary hurdles encountered when employing reinforcement learning in establishing a framework for spatial resource allocation and explore potential research directions that we believe hold promise in addressing these challenges:

1) When using reinforcement learning to optimize spatial resource coverage and facility location selection, there's been limited consideration for the diversity among finite resources in real-world scenarios. For instance, due to budget constraints and inherent scene-related issues, different grades of service facilities might need simultaneous coverage within a single scenario. These facilities exhibit differences in cost, service range, and quality. However, existing studies have uniformly set the attributes of each facility within a scenario, contradicting actual application scenarios. In future reinforcement learning research, it's essential to further refine the differentiation in state descriptions and the hierarchy of service facilities to simulate real-world application scenarios more accurately.

2) Reinforcement learning often models the environment as a grid world to optimize spatial resource allocation and sets the agent's actions to move within the surrounding grid. However, such simulations fail to authentically represent the positional relationship between resource supply points and demand points, as most distances in the real environment are defined by roads. Moreover, natural or artificial boundaries exist, such as grasslands, rivers, and administrative district divisions. When constructing environments for future reinforcement learning, there's a need for more accurate descriptions of distances and various types of boundaries.

3) Considering multiple constraints in the algorithm design of reinforcement learning for solving order matching problems is an important research area. These constraints include user preferences, time windows, cost calculations, structural constraints between pickup and delivery, and more. Effective training in reinforcement learning requires accommodating these additional constraints. While there are existing solutions like tailored designs for these constraints, they are often seen as soft constraints. However, such solutions aren't suitable for scenarios with strict constraints that cannot be easily violated. Modeling these practical challenges effectively remains a key challenge.

4) The core of reinforcement learning training is feedback rewards. It's a valuable research direction in spatial resource allocation to leverage expert data extensively to enhance learning capabilities and save on learning costs. For sparse reward tasks, setting short-term rewards based on the overlap with expert allocation methods can improve learning efficiency.

5) Combining reinforcement learning with graph neural networks (GNN) or transfer learning models (Pan & Yang, 2009) offers a potent approach to addressing spatial resource allocation challenges. GNN demonstrate remarkable capabilities in spatial relationship modeling, effectively capturing the spatial correlations between resources and demand points. They can learn node features, aiding in understanding the attributes of resource and demand points and providing deeper insights for devising resource allocation strategies. Meanwhile, transfer learning facilitates knowledge transfer and model generalization. It enables the application of knowledge learned in one environment to another, assisting models in adapting to new resource allocation scenarios, especially in data-scarce situations. This amalgamation combines the decision optimization prowess of reinforcement learning with the spatial sensitivity of graph neural networks, enabling models to better optimize resource allocation strategies, adapt in real-time adjustments, and learn across diverse environments, thereby enhancing decision-making efficiency and accuracy. Overall, this fusion holds the potential to overcome limitations of reinforcement learning in resource allocation, boosting model performance, adaptability, and generalizability.

## 7 Conclusion

Spatial resource allocation involves distributing finite resources across different locations or regions to meet specific needs or optimize particular objectives. These resources can be various types of facilities, such as charging stations, service centers, transportation nodes, logistics facilities, etc., while demands can arise from daily needs, traffic flows, energy requirements, and more. Key considerations in addressing spatial resource allocation problems encompass the location, quantity, distribution, utilization efficiency of resources, and optimal strategies to fulfill demands. Reinforcement learning, as a potent decision-making and optimization method, offers a new approach and tool for tackling these complex resource allocation issues. In this paper, we conduct a detailed analysis of reinforcement learning methods used in spatial resource allocation problems.

The article divides the spatial resource allocation problem into three categories: static demand-based resource allocation, static resource-based allocation, and dynamic resource allocation. The DQN algorithm, known for solving discrete problems, and its extended algorithms have proven to be the most common and effective approaches for addressing this issue. Both single-agent and multi-agent reinforcement learning algorithms are utilized across various scenarios. Particularly in dynamic resource allocation scenarios, the effectiveness of multi-agent reinforcement learning methods surpasses that of single-agent reinforcement learning algorithms. This is because their flexibility and adaptability enable the system to dynamically adjust resource allocation, making optimal decisions based on evolving conditions, consequently enhancing overall service efficiency. The collaborative decision-making capabilities of multi-agent systems also help in avoiding overlapping resource service areas, promoting information sharing, optimizing vehicle dispatch, and improving system performance.

In different scenarios, the setup of the environment's state, rewards, and actions varies, which is a crucial step in reinforcement learning solutions. Input features are selected based on the objective of the solution and the specific part of the system to achieve the intended goals. Research indicates that solving different parts of the problem might lead to different solutions. Most papers establish objective functions and multiple constraints based on their own scenarios and available data, including them in the algorithm's positive or negative rewards, to generate more practical and utilitarian models.

Based on a detailed review, this paper thoroughly discusses the applicability, advantages, and disadvantages of reinforcement learning in spatial resource allocation. This discussion highlights the key issues in effectively addressing challenges related to spatial resource allocation in the future. Overall, this study offers valuable insights that aid researchers in better understanding the issues, opportunities, and potential directions for the application of reinforcement learning in spatial resource allocation.

**Acknowledgements**

Not applicable.

**Authors' contributions**

Di Zhang, Conceptualization; Investigation; Methodology; Writing original draft; Data curation & Interview. Moyang Wang, Conceptualization; Writing – Review & Editing. Joseph Mango, Conceptualization; Writing – Review & Editing. Xiang Li, Resources; Writing – Review & Editing.

**Funding**

This work was supported by the project funded by International Research Center of Big Data for Sustainable Development Goals [Grant Number CBAS2022GSP07], Chongqing Natural Science Foundation [Grant Number CSTB2022NSCQMSX 2069] and Ministry of Education of China [Grant Number 19JZD023].

**Availability of data and materials**

Not applicable.

**Competing interests**

Not applicable.